\documentclass{article}

\usepackage[final]{nips_2016}

\usepackage[utf8]{inputenc} 
\usepackage[T1]{fontenc}    
\usepackage{url}            
\usepackage{booktabs}       
\usepackage{amsfonts}       
\usepackage{nicefrac}       
\usepackage{microtype}      

\usepackage{amsmath,amssymb,amsthm,amsbsy}
\usepackage{mathrsfs} 
\usepackage{graphicx}	
\usepackage{multirow} 
\usepackage{algorithm} 	
\usepackage{enumerate} 	
\usepackage{caption}
\usepackage{subcaption}

\theoremstyle{plain} 
\theoremstyle{plain} 
\theoremstyle{plain} 
\theoremstyle{definition} 
\theoremstyle{definition} 
\theoremstyle{definition} \newtheorem{assumption}{Assumption}
\theoremstyle{definition} 

\theoremstyle{plain} \newtheorem*{lemma*}{Lemma}
\theoremstyle{plain} \newtheorem*{proposition*}{Proposition}
\theoremstyle{plain} \newtheorem*{theorem*}{Theorem}
\theoremstyle{definition} \newtheorem*{corollary*}{Corollary}
\theoremstyle{definition} \newtheorem*{definition*}{Definition}
\theoremstyle{definition} \newtheorem*{assumption*}{Assumption}
\theoremstyle{definition} \newtheorem*{example*}{Example}

\newcommand{\R}{\ensuremath{\mathbb{R}}}										
 
\newcommand{\Prob}[1]{\mathbb{P}\left[#1\right]}						
\newcommand{\Pb}[2]{\mathbb{P}_{#1} \left[#2\right]}						
\newcommand{\CProb}[2]{\mathbb{P}\left[#1 \left\vert\right. #2 \right]}	

\newcommand{\mb}[1]{\ensuremath{\mathbf{#1}}}
\newcommand{\mc}[1]{\ensuremath{\mathcal{#1}}}							
\newcommand{\wh}[1]{\widehat{#1}}														
\newcommand{\wt}[1]{\widetilde{#1}}													

\newcommand{\beq}[1]{\begin{equation} \label{eq:#1}}
\newcommand{\eeq}{\end{equation}}
\newcommand{\beqn}{\begin{equation*}}
\newcommand{\eeqn}{\end{equation*}}

\title{A nonparametric sequential test for online randomized experiments}

\author{
  Vineet Abhishek\\
  Walmart Labs\\
  CA, USA\\
  \texttt{vabhishek@walmartlabs.com} \\
  \And
  Shie Mannor\\
  Technion\\
  Haifa, Israel\\
  \texttt{shie@ee.technion.ac.il} \\
}

\begin{document}

\maketitle

\begin{abstract}
We propose a nonparametric sequential test that aims to address two practical problems pertinent to online randomized experiments: (i) how to do a hypothesis test for complex metrics; (ii) how to prevent type $1$ error inflation under continuous monitoring. The proposed test does not require knowledge of the underlying probability distribution generating the data. We use the bootstrap to estimate the likelihood for blocks of data followed by mixture sequential probability ratio test. We validate this procedure on data from a major online e-commerce website. We show that the proposed test controls type $1$ error at any time, has good power, is robust to misspecification in the distribution generating the data, and allows quick inference in online randomized experiments.
\end{abstract}

\section{Introduction} \label{sec:intro}
Most major online websites continuously run several randomized experiments to measure the performance of new features or algorithms and to determine their successes or failures; see \cite{kohavi-2009} for a comprehensive survey. A typical workflow consists of: (i) defining a success metric; (ii) performing a statistical hypothesis test to determine whether or not the observed change in the success metric is significant. Both of these steps, however, are nontrivial in practice. 

Consider a typical product search performance data collected by any online e-commerce website. Such data is multidimensional, containing dimensions such as the number of search queries, the number of conversions, time to the first click, total revenue from the sale, etc., for each user session. The success metric of interest could be a complicated scalar function of this multidimensional data rather than some simple average; e.g., a metric of interest could be the ratio of expected number of successful queries per user session to the expected number of total queries per user session, an empirical estimate of which is a ratio of two sums. Furthermore, some commonly used families of probability distributions such as the Bernoulli distribution or the normal distribution are almost never suitable to model this data. The search queries in a user session are highly correlated; using the Bernoulli distribution to model even simple metrics such the click through rate per impression is inaccurate.\footnote{A similar observation is made by \cite{crook-2009}. The authors suggest avoiding the use of standard formulas for standard deviation computation which assume independence of the samples and recommend the use of the bootstrap.} Another example is the revenue based metrics with a considerable mass at zero and a heavy tail distribution for the nonzero values. The use of the normal distribution for modeling revenue per user session is inaccurate and sensitive to extreme values unless the sample size is huge. As a result, the use of the binomial proportion test or the t-test for hypothesis testing is unsuitable for such metrics respectively. It requires considerable effort to find appropriate models for each of these metrics and it is implausible to come up with a suitable parametrized family of probability distributions that could be used to model different complex success metrics. 

Additionally, classical statistical tests such as the t-test assume that the number of samples is given apriori. This assumption easily breaks in an online setting where an end user with not much formal training in statistics has near real-time access to the performance of ongoing randomized experiments: the decision to stop or continue collecting more data is often based on whether or not the statistical test on the data collected so far shows significance or not. This practice of chasing significance leads to highly inflated type~$1$ error.\footnote{The inflation of type $1$ error because of data collection to reach significance can be explained by the law of the iterated logarithm, as observed in \cite{pekelis-2015}. Section \ref{sec:eval} provides some plots to demonstrate the ill effects of chasing significance.} Several recent articles have highlighted this aspect of data collection as one of the factors contributing to the statistical crisis in science; see, e.g., \cite{simmons:2011} and \cite{goodson-2014}. Use of sample size calculator to determine the number of samples to be collected and waiting until the required number of samples have been collected is hard to enforce in practice, especially in do-it-yourself type platforms commonly used to set up and monitor online randomized experiments. Additionally, the sample size calculation requires an estimate of the minimum effect size and the variance of the metric. These estimates are often crude and result in waiting longer than necessary in case the true change is much larger than the estimate.

The above two challenges point to two separate fields of statistics: the bootstrap and sequential tests. The bootstrap is frequently used to estimate the properties of complicated statistics; see \cite{efron-1993} and \cite{hall-1997} for a comprehensive review. Sequential tests provide an alternative to committing to a fixed sample size by providing type~$1$ error control at any time under continuous monitoring; see \cite{Lai-2001} for a survey on sequential tests. Sequential tests, however, are still uncommon in practice partly because of the information they require about the data generation process.

Our main contribution is in bringing together the ideas from the bootstrap and sequential tests and applying them to construct a nonparametric sequential test for performing a hypothesis test on complex metrics. We provide empirical evidence using the data from a major online e-commerce website demonstrating that our nonparametric sequential test is suitable for the diverse type of metrics, ensures that type~$1$ error is controlled at any time under continuous monitoring, has good power, is robust to misspecification in the distribution generating the data, and enables making quick inference in online randomized experiments. We advocate the use of this general approach in the absence of knowledge of the right distribution generating the data, which typically is the case. 

Briefly, we compute a studentized statistic for the metric of interest on blocks of data and use the bootstrap to estimate its likelihood in a nonparametric manner. The likelihood ratios can then be used to perform mixture sequential probability ratio test (henceforth, mixture SPRT) \cite{robbins-1970}. We highlight theoretical results from existing literature justifying the use of the bootstrap based likelihood estimation and the mixture SPRT.

\textbf{Related work}: The bootstrap based nonparametric approach we take distinguishes our work from the related literature on sequential testing. Most of the existing literature assumes the knowledge of parametrized family of distribution from which data (typically scalar) is generated; ours does not. The idea of sequential probability ratio test (SPRT) for testing a simple hypothesis and its optimality originates from the seminal work of \cite{wald-1945}. Mixture SPRT \cite{robbins-1970}, described in Section \ref{sec:m-sprt}, extends this to a composite hypothesis. For the exponential family of distributions, mixture SPRT is shown to be a test with power $1$ and almost optimal with respect to expected time to stop; see \cite{robbins-1970} and \cite{pollak-1978}. \cite{johari-2016} formalizes the notion of always valid p-value and uses mixture SPRT for the sequential test. MaxSPRT \cite{kulldorff}, \cite{kharitonov-2015} is another popular sequential test where the unknown parameter is replaced by its maximum likelihood estimate computed from the observed data so far and the rejection threshold is determined (typically using numerical simulations) to ensure type $1$ error control. The computation of the maximum likelihood estimate as well as the rejection threshold requires knowing the parametrized family of distribution; consequently, MaxSPRT has been studied only for a few simple distributions. Group sequential tests \cite{pocock-1977} divide observations into blocks of data for inference. It, however, assumes the normal distribution and uses mean as the metric of interest. The theoretical foundations of the bootstrap likelihood estimation can be found in \cite{hall-1987}; it shows that the bootstrap based likelihood estimation of the studentized statistic provide a fast rate of convergence to the true likelihood. \cite{davison-1992} uses a nested bootstrap to compute the likelihood which can be computationally expensive. \cite{efron-1993b} uses confidence intervals to compute the likelihood and shows that for the exponential family of distributions, the confidence interval based approach has the rate of convergence of the same order as in \cite{hall-1987}.

\section{Model, notation, and assumptions} \label{sec:model}
Let $\mb{X} \in \R^d$ be a random vector with the CDF $F(\mb{x}; \theta, \eta)$ and corresponding pdf $f(\mb{x}; \theta, \eta)$. Here $\theta$ is the scalar parameter of interest and $\eta$ is a nuisance parameter (possibly null or multidimensional). Denote a realized value of the random vector $\mb{X}$ by $\mb{x}$. For $j < k$, let $\mb{X}_{j:k} \triangleq (\mb{X}_j, \mb{X}_{j+1}, \ldots, \mb{X}_{k})$ be a collection of i.i.d. random vectors indexed from $j$ to $k$; define $\mb{x}_{j:k}$ similarly. 

We are interested in testing the null hypothesis that $\theta = \theta_0$ against the alternate hypothesis $\theta \neq \theta_0$ such that type 1 error is controlled at any given time and the power of the test should be high; i.e., we are looking for a function $D_{\theta_0}$ mapping sample paths $\mb{x}_1, \mb{x}_2, \ldots $ to $[0, 1]$ such that for any $\alpha \in (0, 1)$:
\beq{valid-pvalue}
\Pb{\theta}{D_{\theta_0}(\mb{X}_{1:n}) \leq \alpha ~ \text{for some $n \geq 1$} } \quad \left\{
\begin{array}{l l}
	\leq \alpha & \quad \text{if $\theta = \theta_0$,}\\
	\approx 1 & \quad \text{if $\theta \neq \theta_0$,}
\end{array} \right.
\eeq
where $\mb{X}$'s are assumed to be realized i.i.d. from $F(\mb{x}; \theta, \eta)$ and $\mathbb{P}_{\theta}$ is the probability under $F(\mb{x}; \theta, \eta)$. The first part of \eqref{eq:valid-pvalue} is a statement on type $1$ error control: if the null hypothesis is true and our tolerance for incorrect rejection of the null hypothesis is $\alpha$, then the probability that the test statistic exceeds $\alpha$ and rejects the null hypothesis (\textit{type $1$ error}) is at most $\alpha$. The second part of \eqref{eq:valid-pvalue} is a statement on the power of the test: if the null hypothesis is false then the probability that the test statistic exceeds $\alpha$ and rejects the null hypothesis (\textit{the power of the test}) is close to one.

Let $\wh{F}(\mb{x}_{1:n})$ be the empirical CDF constructed from $n$ i.i.d. samples. We make the following assumption about the parameter of interest $\theta$:

\begin{assumption} \label{assume:normal}
Assume that $\theta$ is a functional statistic with asymptotically normal distribution, i.e., $\theta = T(F)$ for some functional $T$ and $\sqrt{n}(T(\wh{F}(\mb{X}_{1:n})) - \theta) \rightarrow \mc{N}(0, \zeta^2)$ in distribution as $n \rightarrow \infty$, where $T(\wh{F}(\mb{x}_{1:n}))$ is the plug-in estimate of $\theta$. Furthermore, assume that the asymptotic variance $\zeta^2$ does not depend on the specific choice of $\theta$.
\end{assumption}

Asymptotic normality of plug-in statistic in Assumption~\ref{assume:normal} holds under fairly general conditions and captures several statistics of interest; see \cite{lehmann-2004} for a formal discussion on this.

\section{Preliminaries} \label{sec:preliminaries}
Our work leverages ideas from the bootstrap and mixture SPRT. This section provides their brief overview. 

\subsection{The bootstrap} \label{sec:bs}
Let $Z_{1:n}$ be $n$ i.i.d. random variables and let $H_n(z) \triangleq \Prob{T(Z_{1:n}) \leq z}$ be the sampling distribution of a functional $T$. The CDF of $Z$'s is usually unknown and hence $H_n$ cannot be computed directly. The (nonparametric) bootstrap is used to approximate the sampling distribution $H_n$ using a resampling technique. Let $Z^*_{1:n}$ be~$n$ i.i.d. random variables each with the CDF equal to the empirical CDF constructed from $Z_{1:n}$. Let $H^*_n(z) \triangleq \CProb{T(Z^*_{1:n}) \leq z}{Z_{1:n}}$ be the bootstrap distribution of the functional $T$. Notice that $H_n^*$ is a random CDF that depends on the realized values of $Z_{1:n}$. The idea is to use $H_n^*$ as an approximation for  $H_n$.\footnote{Under some regularity conditions and a suitable distance metric on the space of CDFs, distance between $H_n$ and $H_n^*$ goes to zero with probability $1$ as $n \rightarrow \infty$; see \cite{hall-1997}.} 

The bootstrap distribution is computed using Monte Carlo method by drawing a large number of samples, each of size $n$, from the empirical distribution. This is identical to sampling with replacement and provides a convenient computational method to deal with complex statistics. Even in the cases where the statistics are asymptotically normal, we might have to rely on the bootstrap to compute its asymptotic variance. Moreover, for asymptotically normal statistics, the bootstrap typically provides an order improvement in convergence rate to the true distribution (see \cite{hall-1997} and \cite{hall-1987}). This motivates the use of the bootstrap in our nonparametric sequential test.

\subsection{Mixture SPRT} \label{sec:m-sprt}
In a general form, mixture SRPT works as follows. Let $\rho_n(z_{1:n}; \theta)$ be the joint pdf parametrized by~$\theta$ of a sequence of random variables $Z_{1:n}$. Let $\theta \in \mc{I}$ and $\pi(\theta)$ be any prior on $\theta$ with $\pi(\theta) > 0$ for $\theta \in \mc{I}$. Define:
\beq{msprt}
L(z_{1:n};\theta_0) \triangleq \frac{\int_{\theta \in \mc{I}} \rho_n(z_{1:n}; \theta) \pi(\theta)}{ \rho_n(z_{1:n}; \theta_0)}.
\eeq
If $\theta = \theta_0$, the sequence $L(Z_{1:n};\theta_0)$'s forms a martingale with respect to the filtration induced by the sequence $Z_n$'s.\footnote{This follows from a general observation that if $Z_{1:n} \sim \rho_n(z_{1:n})$ and $\psi_n(z_{1:n})$ is any other probability density function, then $L_n = \nicefrac{\psi_n(z_{1:n})}{\rho_n(z_{1:n})}$ is a martingale.} Using Doob's martingale inequality, 
\beq{doob}
\Pb{\theta_0}{L(Z_{1:n};\theta_0) > \frac{1}{\alpha} ~ \text{for some $n \geq 1$}} \leq \alpha \text{ for any } \alpha > 0, 
\eeq
where the probability is computed assuming $\theta = \theta_0$. An immediate consequence of this is control of type~$1$ error at any time. The optimality of mixture SPRT with respect of power and the expected time to stop (\cite{robbins-1970} and \cite{pollak-1978}) and its ease of implementation motivates its use in our nonparametric sequential test.

The p-value at time $n$ is defined as:
\beq{p-val}
p_{value}(z_{1:n};\theta_0) \triangleq \min\left\{ 1, \frac{1}{\max\{L(z_{1:t}; \theta_0): t \leq n\}} \right\}.
\eeq
It follows immediate from \eqref{eq:doob} and \eqref{eq:p-val} that 
\beq{p-val-valid}
\Pb{\theta_0}{p_{value}(Z_{1:n};\theta_0) \leq \alpha} \leq \alpha \quad \text{for any $n$}
\eeq
and $p_{value}(z_{1:n};\theta_0)$ is nonincreasing in $n$. In this sense, this is an always valid p-value sequence, defined in \cite{johari-2016}. The duality between the p-value and the confidence interval can be used to construct an always valid (though possibly larger than the exact) confidence interval.

Mixture SPRT, however, requires the knowledge of the functional form of the pdf up to the unknown parameter of interest. This is hard especially in a multidimensional setting and in the presence of nuisance parameter(s). We address this using the bootstrap, as described next.

\section{The nonparametric sequential test} \label{sec:seq-test}
This section shows how mixture SPRT and the bootstrap could be used together to construct a nonparametric sequential test for complex statistics. We will also discuss how to apply the ideas developed here to A/B tests. 

\subsection{The bootstrap mixture SPRT} \label{sec:bs-seq-test}
Our main idea is as follows: (i) divide the data into blocks of size $N$; (ii) compute a studentized plug-in statistic for $\theta$ on each block of data; (iii) estimate the likelihood of the studentized plug-in statistic for each block using the bootstrap; (iv) use mixture SPRT by treating each block as a unit observation. We provide details below.

Let $\mb{x}_{(k)} \triangleq \mb{x}_{kN:(k+1)N-1}$ denote the $k^{th}$ block of data of size $N$ and $s(\mb{x}_{(k)}; \theta)$ be the studentized plug-in statistic computed from the $k^{th}$ data block, i.e.,
\beq{student}
s(\mb{x}^{(k)}; \theta) = \frac{T(\wh{F}(\mb{x}_{(k)})) - \theta}{\sigma(T(\wh{F}(\mb{x}_{(k)})))},
\eeq
where $\sigma(T(\wh{F}(\mb{x}_{(k})))$ is a consistent estimate of the standard deviation of the plug-in statistic $T(\wh{F}(\mb{X}_{(k)}))$.\footnote{Formally, $\sqrt{n}\sigma(T(\wh{F}(\mb{X}_{1:n}))) \rightarrow \zeta$ with probability $1$ as $n \rightarrow \infty$, where $\zeta$ is defined in Assumption \ref{assume:normal}.} For notational convenience, we use $\wh{\theta}_{(k)}$ and $s_{(k)}^{\theta}$ in lieu of $T(\wh{F}(\mb{x}_{(k)}))$ and $s(\mb{x}_{(k)};\theta)$ respectively, with the dependence on underlying $\mb{x}$'s and $N$ being implicit; and their uppercase versions to denote the corresponding random variables. The random variable $S_{(k)}^{\theta}$ is asymptotically distributed as $\mc{N}(0, 1)$ at rate $O(\nicefrac{1}{\sqrt{N}})$ assuming $\theta$ is the true parameter. Hence for large $N$, the distribution of $S_{(k)}^{\theta}$ is approximately pivotal (i.e., the distribution does not depend on the unknown $\theta$ and the effect of $\eta$ is negligible). This suggests using the normal density as the pdf for $S_{(k)}^{\theta}$ and carry out mixture SPRT following the steps in Section \ref{sec:m-sprt}. The density, however, can be approximated better using the bootstrap.

Algorithm \ref{algo:bs-sprt} below describes the bootstrap mixture SPRT.

\begin{algorithm}
\floatname{algorithm}{Algorithm}
\caption{The bootstrap mixture SPRT} \label{algo:bs-sprt}
Given blocks of observations $\mb{x}_{(k)}$ for $k = 1, 2, \ldots$, and a prior $\pi(\theta)$: 
\begin{enumerate}
\item
Draw $\wt{\theta}_1, \wt{\theta}_2, \ldots \wt{\theta}_M$ i.i.d. from $\pi(\theta)$ for some large $M$.
\item
For each $k$:
\begin{enumerate}
\item
Compute $\wh{\theta}_{(k)}$ and its standard deviation $\sigma(\wh{\theta}_{(k)})$.
\item
Draw a random sample of size $N$ with replacement from $\mb{x}_{(k)}$; denote it by $\mb{x}^*_{(k)}$. Repeat this $B$ times to obtain a sequence $\mb{x}_{(k)}^{*b}$, $1 \leq b \leq B$. For each b, compute the studentized plug-in statistic $s^{*b}_{(k)}$ as:
\beq{student-bs}
s^{*b}_{(k)} = \frac{T(\wh{F}(\mb{x}_{(k)}^{*b})) - \wh{\theta}_{(k)}}{\sigma(T(\wh{F}(\mb{x}_{(k)}^{*b})))}.
\eeq
Let $g^*_{(k)}(\cdot)$ be the pdf obtained using a Gaussian kernel density estimate from the sequence $s^{*b}_{(k)}$, $1 \leq b \leq B$.
\end{enumerate}
\item
Compute $L_n$ as follows:
\beq{likelihood-bs}
L_n = \frac{1}{M}\left(\sum_{m=1}^M\left(\frac{\prod_{k=1}^n g^*_{(k)}\left(\frac{\wh{\theta}_{(k)} - \wt{\theta}_m}{\sigma(\wh{\theta}_{(k)})}\right) }{\prod_{k=1}^n g^*_{(k)}\left(\frac{\wh{\theta}_{(k)} - \theta_0}{\sigma(\wh{\theta}_{(k)})}\right)} \right)\right.
\eeq
\item
Reject the null hypothesis $\theta = \theta_0$ if $L_n > 1/\alpha$ for some $n > 1$.
\end{enumerate}
\end{algorithm}

\textbf{Correctness of Algorithm \ref{algo:bs-sprt}}: We give a heuristic explanation of the correctness of Algorithm~\ref{algo:bs-sprt}; a rigorous proof of type $1$ error control or optimality of Algorithm \ref{algo:bs-sprt} in some sense is extremely hard.\footnote{The results on the optimality of mixture SPRT are known only for exponential families of distributions. We, however, take a nonparametric approach. A theory of bootstrap is based on Edgeworth expansion \cite{hall-1997}. Applying Edgeworth expansion to analyze likelihood ratio statistic defined in \eqref{eq:likelihood-bs} is quite challenging.} $L_n$ in step $3$ is similar to \eqref{eq:msprt}. If $L_n$ is approximately a martingale for $\theta = \theta_0$, the arguments of Section~\ref{sec:m-sprt} would imply type $1$ error control at any time under continuous monitoring. Step $2$ of Algorithm~\ref{algo:bs-sprt} is approximating the true pdf of the random variable $S_{(k)}^{\theta}$, denoted by $g^{\theta}_{(k)}$, using the bootstrap based estimate $g^*_{(k)}$. We only need to ensure that $g^*_{(k)}$ is a good approximation of $g^{\theta_0}_{(k)}$, assuming $\theta = \theta_0$, for the martingale property to hold. A subtlety arises because the bootstrap distribution is discrete. This, however, is not a problem: \cite{hall-1987} shows that a smoothened version of bootstrap density obtained by adding a small independent noise $\mc{N}(0, \nicefrac{1}{N^c})$ with arbitrary large $c$ to the bootstrap samples converges uniformly to the true density at rate $O_p(\nicefrac{1}{N})$. Working with the studentized statistic is crucial in obtaining $O_p(\nicefrac{1}{N})$ convergence; see \cite{hall-1997} for further details. By contrast, approximating $g_{(k)}$ by the normal density is accurate to $O(\nicefrac{1}{\sqrt{N}})$. The bootstrap samples behave as if realized from a true density. A Gaussian kernel density can be used to estimate the bootstrap density from a finite number of bootstrap samples, as in step $2$ of Algorithm~\ref{algo:bs-sprt}. The independence of the asymptotic variance and $\theta$ by Assumption \ref{assume:normal} allows us to compute the likelihood for different values of $\theta$. Step $4$ is a Monte Carlo computation of Bayesian average in the right side of \eqref{eq:msprt}. 

\subsection{Application to A/B tests} \label{sec:abtest}
In A/B tests, the goal is to identify whether or not the success metric computed on each of the two groups of data has a nonzero difference (i.e., $\theta_0 = 0$). The bootstrap mixture SPRT easily extends to A/B tests. Assume for simplicity that the block size is the same for both groups. Let $\mb{x}_{(k)}$'s be data from group $A$ (control) and $\mb{y}_{(k)}$'s the data from group $B$ (variation). We modify the quantities $\wh{\theta}_{(k)}$, $\sigma(\theta_{(k)})$, and $s^{*b}_{(k)}$ of Algorithm \ref{algo:bs-sprt} as follows:
\begin{align*}
\wh{\theta}_{(k)} &= T(\wh{F}(\mb{y}_{(k)})) - T(\wh{F}(\mb{x}_{(k)})), \\
\sigma(\wh{\theta}_{(k)}) &= \sqrt{\sigma^2(T(\wh{F}(\mb{y}_{(k)}))) + \sigma^2(T(\wh{F}(\mb{x}_{(k)})))}, \\
s^{*b}_{(k)} &= \frac{T(\wh{F}(\mb{y}_{(k)}^{*b})) - T(\wh{F}(\mb{x}_{(k)}^{*b})) - \wh{\theta}_{(k)}}{\sqrt{\sigma^2(T(\wh{F}(\mb{x}_{(k)}^{*b}))) + \sigma^2(T(\wh{F}(\mb{x}_{(k)}^{*b})))}}.
\end{align*}

We conclude with a discussion on how to determine various parameters in Algorithm \ref{algo:bs-sprt}. Type~$1$ error control holds for any prior $\pi(\theta)$ with nonzero density on possible values of $\theta$. The choice of prior, however,  can affect the number of observations needed to reject the null hypothesis. A simple choice is $\mc{N}(0, \tau^2)$; $\tau$ can be determined from the outcomes of the past A/B tests or can be set to a few percent of the reference value, computed from the data collected before the start of the A/B test, of the metric of interest. The standard deviation in \eqref{eq:student-bs} for each $b$ can be computed using the bootstrap again on $\mb{x}^{*b}_{(k)}$. This, however, is computationally prohibitive. In practice, the standard deviation is computed using the delta method or the jackknife; see e.g., \cite{efron-1993}. A similar method can be used to compute the standard deviation in \eqref{eq:student}. The parameters $B$ and $M$ are merely used for Monte Carlo integration; any sufficiently large value is good. Typical choices for the number of bootstrap sample $B$ is $1000-5000$ and the number of Monte Carlo samples $M$ is $5000-10,000$. Because we work with a block of data at a time, the computational overhead is not much; computation can be performed in an online manner as new blocks of data arrive or using a MapReduce approach on the historical data. Finally, a small block size is intuitively expected to enable fast inference. Doob's martingale inequality is far from being tight for martingales stopped at some maximum length and becomes tighter as this maximum length is increased. However, the small block size may reduce the accuracy of the bootstrap likelihood estimate. Additionally, data update frequency of a near real-time system puts a lower limit on how small the block of data can be; it is not uncommon to receive a few hundred or a few thousand data points per update. Section \ref{sec:eval} describes how to estimate a reasonable block size using post A/A tests.

\section{Empirical evaluation and discussion} \label{sec:eval}

\textbf{Datasets, metrics, and baselines}: We use data obtained from a major online e-commerce website on product search performance. Each dataset contains a tiny fraction of randomly chosen anonymized users; and for each user session, it contains the session timestamp, the number of distinct search queries, the number of successful search queries, and the total revenue. Datasets contain about 1.8 million user sessions each and span nonoverlapping 2-3 weeks timeframe. We show here the results for only one dataset; similar results were observed for the other datasets. 

We work with two metrics: (i) \textit{query success rate per user session}; (ii) \textit{revenue per user session}. The former is the ratio of the sum of successful queries per user session to the sum of total queries per user session, where the sum is over user sessions. The latter is a simple average, however, with most values being zero and nonzero values showing a heavy tail. These two very different metrics are chosen to evaluate the flexibility of the bootstrap mixture SPRT. We have used $\mc{N}(0, \tau^2)$ distribution as prior $\pi(\theta)$, with $\tau$ being equal to a few percent (less than 5\%) of the values of metrics computed on data before the start timestamp of the dataset under consideration. Data is partitioned into blocks based on the order induced by the user session timestamp.

Existing sequential tests require knowing a parameterized distribution generating the data. The data generation model for the two metrics we study in our experiments is hard to find. We are not aware of any existing sequential tests that can be applied to the two complex metrics we use in our experiments. Consequently, we use the z-test on the entire data as a baseline for comparison.\footnote{Because of the large size of the dataset we use, there is no observable difference between the t-test and the z-test.} The use of z-test (or the t-test) is pervasive in commercial A/B tests platforms, often without much scrutiny. Additionally, we compare the bootstrap mixture SPRT with MaxSPRT (\cite{kulldorff}, \cite{kharitonov-2015}) on synthetic data generated from the Bernoulli distribution. MaxSPRT requires knowing a parametrized likelihood function and has been applied mostly when the data is generated from simple distributions such as the Poisson distribution and the Bernoulli distribution.

\textbf{Post A/A tests based methodology}: We make use of post A/A tests to estimate the type $1$ error. For each dataset, we randomly divide the users into two groups, perform the statistical test for the null hypothesis of zero difference in the metrics for the two groups, and compute the p-values. This process is repeated multiple times ($250$ in our experiments). Because there is no true difference between the two groups, the p-value should theoretically follow uniform distribution between $0$ and~$1$. Consequently, the Q-Q plots for the realized p-values from different random splits and the uniform distribution should be the line of slope~$1$. To estimate the power of the test, for each random split of the dataset, we add to the realized difference in the metrics for the two groups some small positive offset. For each choice of the additive offset, we then check how many times the null hypothesis is rejected using p-value $\leq 0.05$ as our null hypothesis rejection criteria.

\begin{figure}[t] 
\begin{center}
\begin{subfigure}[t]{1.0\textwidth}
\centering
\includegraphics[trim=0.7in 0.0in 0.95in 0.25in, clip=true, height=2.2in]{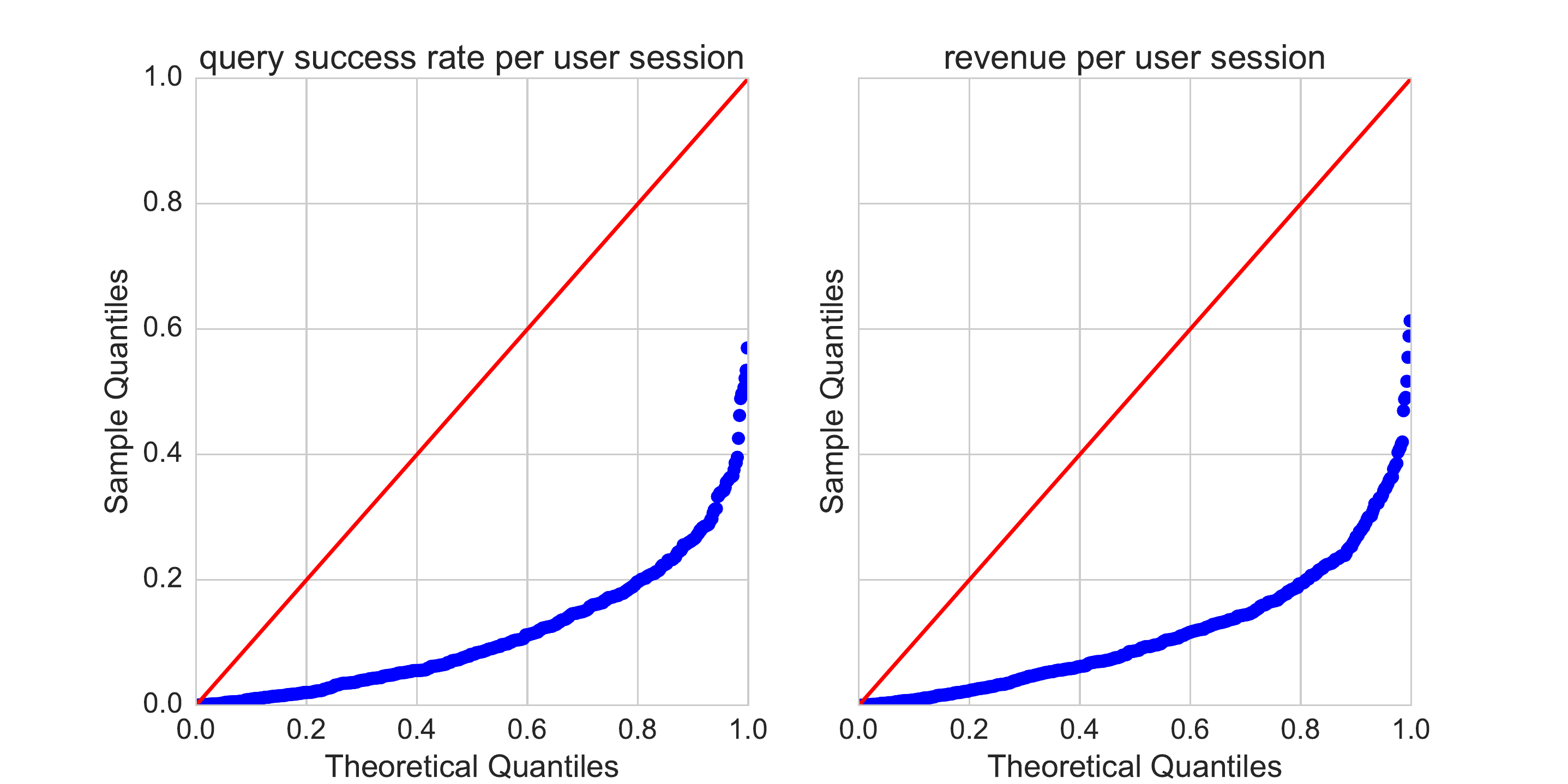}
\caption{\small \sl p-values under sequential observation for the z-test.\label{fig:ztest-daily}} 
\end{subfigure}
~~
\begin{subfigure}[t] {1.0\textwidth}
\centering
\includegraphics[trim=0.7in 0.0in 0.95in 0.25in, clip=true, height=2.2in]{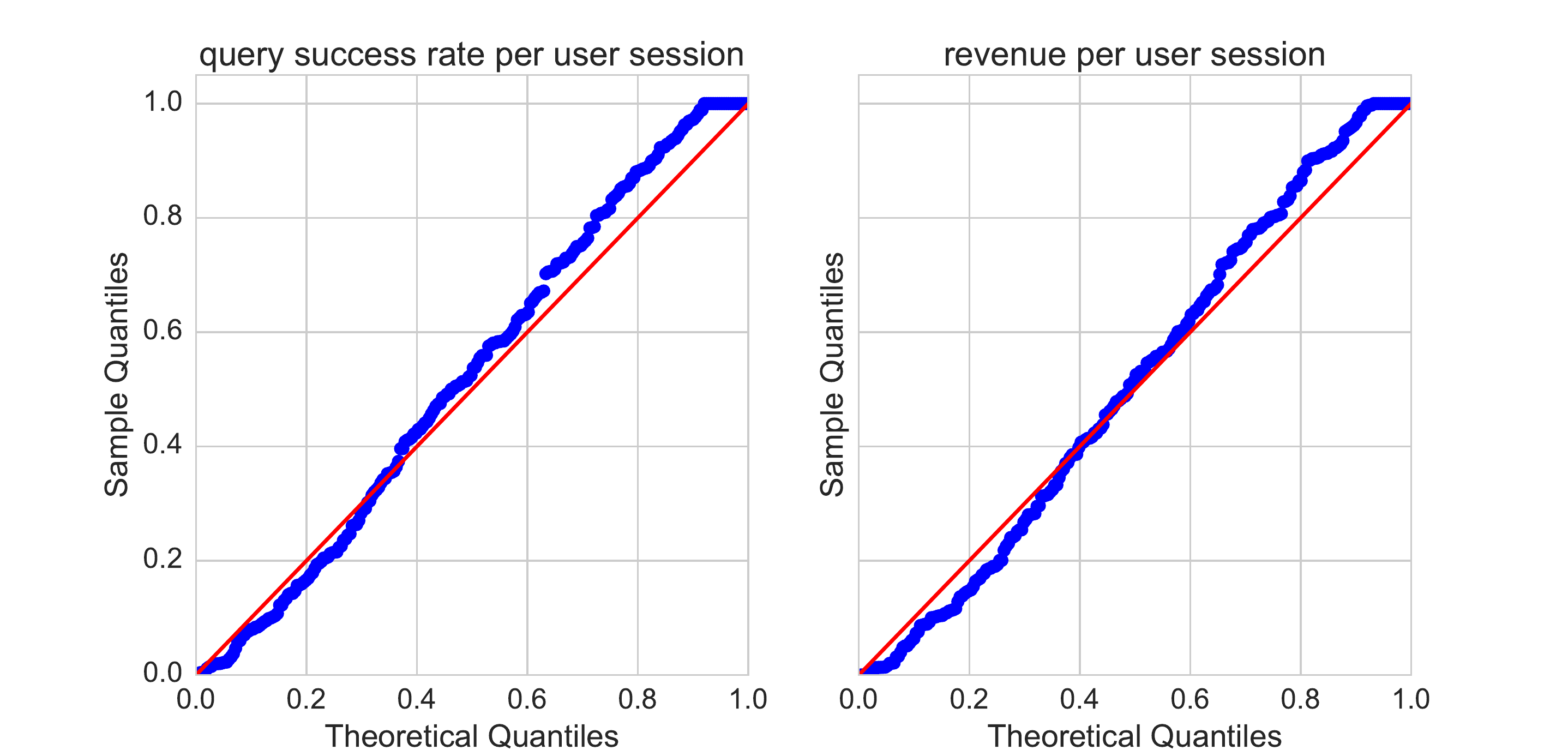}
\caption{\small \sl p-values under the bootstrap mixture SPRT.\label{fig:aatest-5k}} 
\end{subfigure}
\end{center}
\caption{Q-Q plot of p-values computed on random splits of the dataset.}
\end{figure}

\textbf{Type 1 error}: Figure \ref{fig:ztest-daily} shows a huge inflation of type $1$ error under sequential monitoring in case of classical fixed sample size z-test. The Q-Q plot lies well below the line of slope~$1$. Here, the p-values are computed daily using the data collected until that day. The Q-Q plot is for the minimum p-value observed from daily observations, modeling the scenario where one waits until the data shows significance. Waiting until the p-value is $\leq 0.05$ and stopping the test on that day resulted in type $1$ error rate of at least $30\%$. By contrast, the Q-Q plot of always valid p-value in the case of the bootstrap mixture SPRT with the block size $5000$ is close to the line of slope~$1$, as shown in Figure \ref{fig:aatest-5k}. This implies that the bootstrap mixture SPRT controls type $1$ error at any time under continuous monitoring. The p-value is the final one at the end of all the blocks (recall that the p-value is nonincreasing in this case). 

\begin{figure}[t] 
\begin{center}
\begin{subfigure}[t]{1.0\textwidth}
\centering
\includegraphics[trim=0.7in 0.0in 0.95in 0.25in, clip=true, height=2.2in]{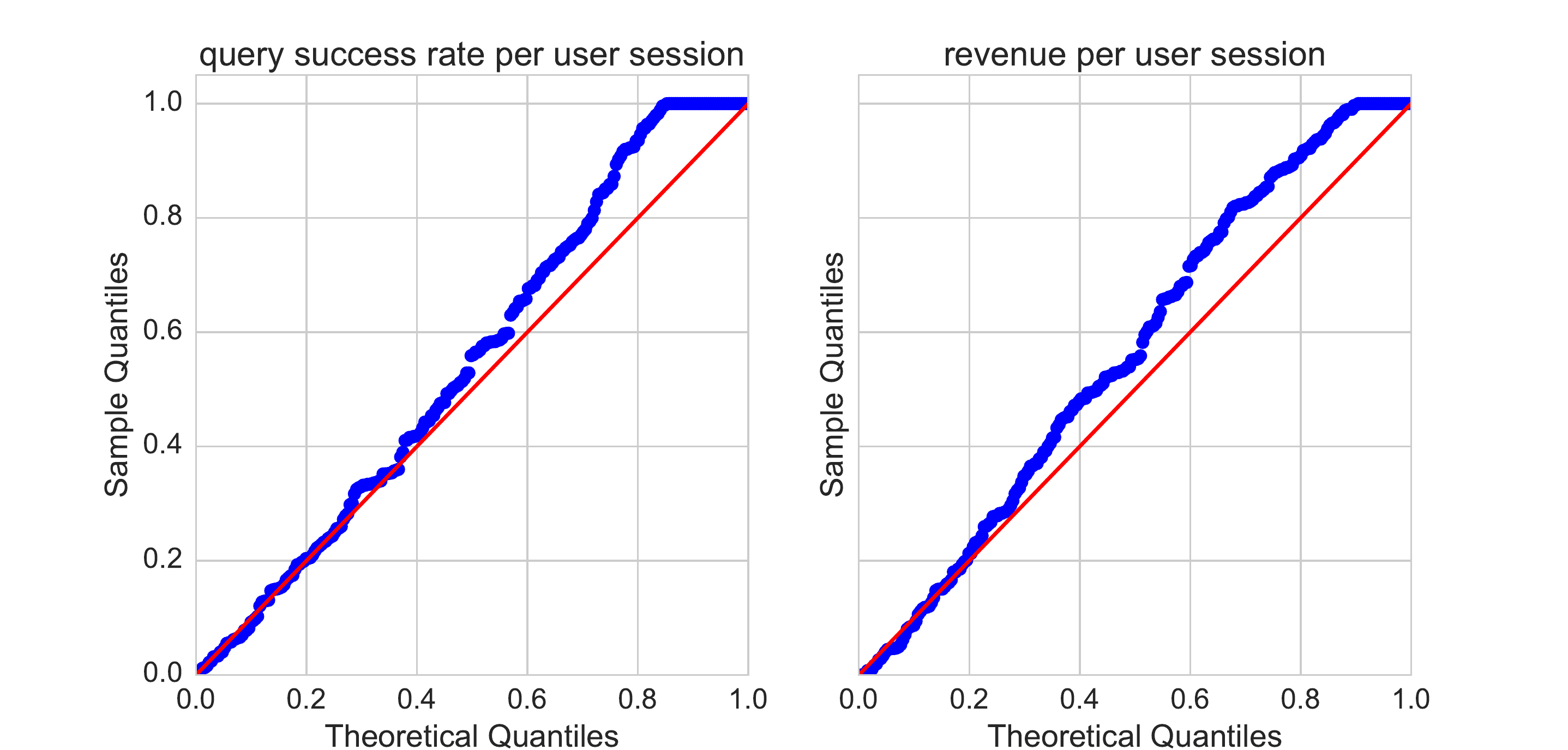}
\caption{\small \sl p-values for block size 10,000.\label{fig:aatest-10k}} 
\end{subfigure}
~~
\begin{subfigure}[t]{1.0\textwidth}
\centering
\includegraphics[trim=0.7in 0.0in 0.95in 0.25in, clip=true, height=2.2in]{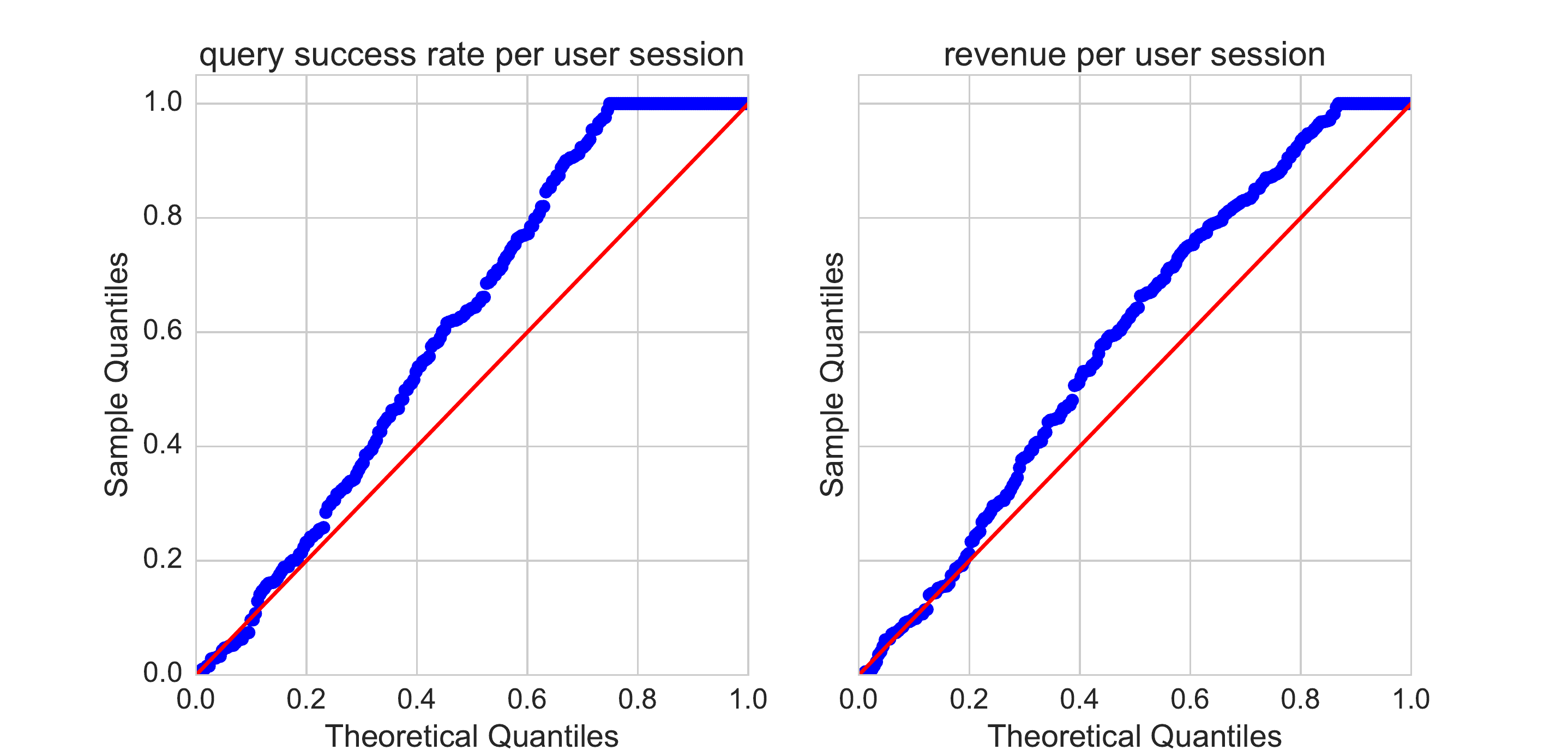}
\caption{\small \sl p-values for block size 20,000.\label{fig:aatest-20k}} 
\end{subfigure}
\end{center}
\caption{Q-Q plot of p-values under bootstrap mixture SPRT with different block sizes.}
\end{figure}

The choice of the block size of $5000$ is determined by post A/A tests with different block sizes, looking at the corresponding Q-Q plots, and making sure it stays above but close to the line of slope~$1$. Smaller block size makes the bootstrap approximation inaccurate while the larger block size makes this procedure more conservative in case of some upper limit on the number of samples available for the test. This is a consequence of fewer intermediate observations in a finite length dataset, resulting in Doob's martingale inequality far from being tight. Figure \ref{fig:aatest-10k} and \ref{fig:aatest-20k} show the Q-Q plot for block length of $10,000$ and $20,000$ respectively. The plots lie above the line of slope~$1$, showing a more conservative type~$1$ error control.

\begin{figure}[ht] 
\begin{center}
\begin{subfigure}[t]{0.5\textwidth}
\centering
\includegraphics[trim=0.25in 0.0in 0.65in 0.25in, clip=true, height=2in]{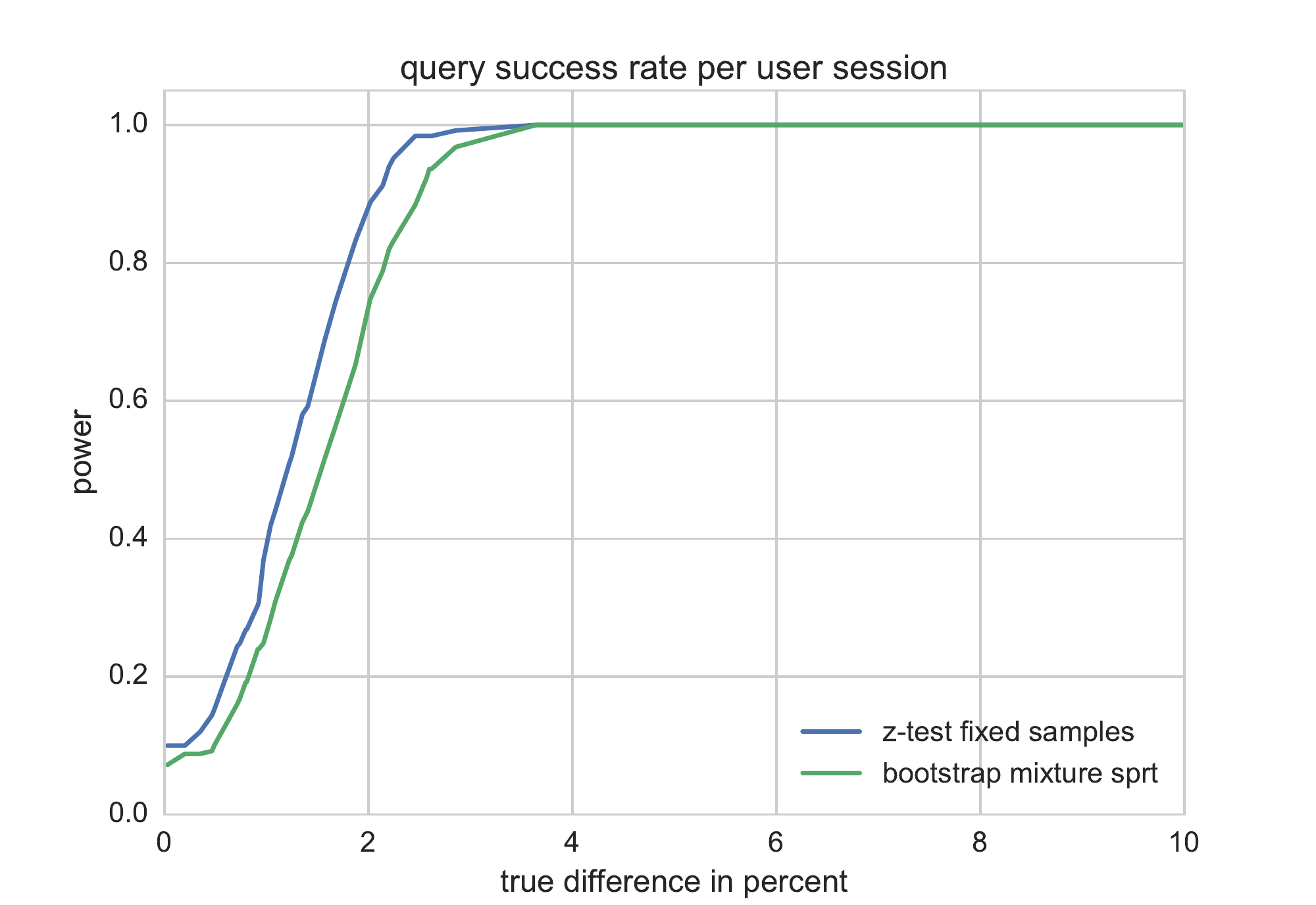}
\caption{\small \sl test power for query success rate per user session.\label{fig:atc-power}} 
\end{subfigure}~~
\begin{subfigure}[t] {0.5\textwidth}
\centering
\includegraphics[trim=0.25in 0.0in 0.65in 0.25in, clip=true, height=2in]{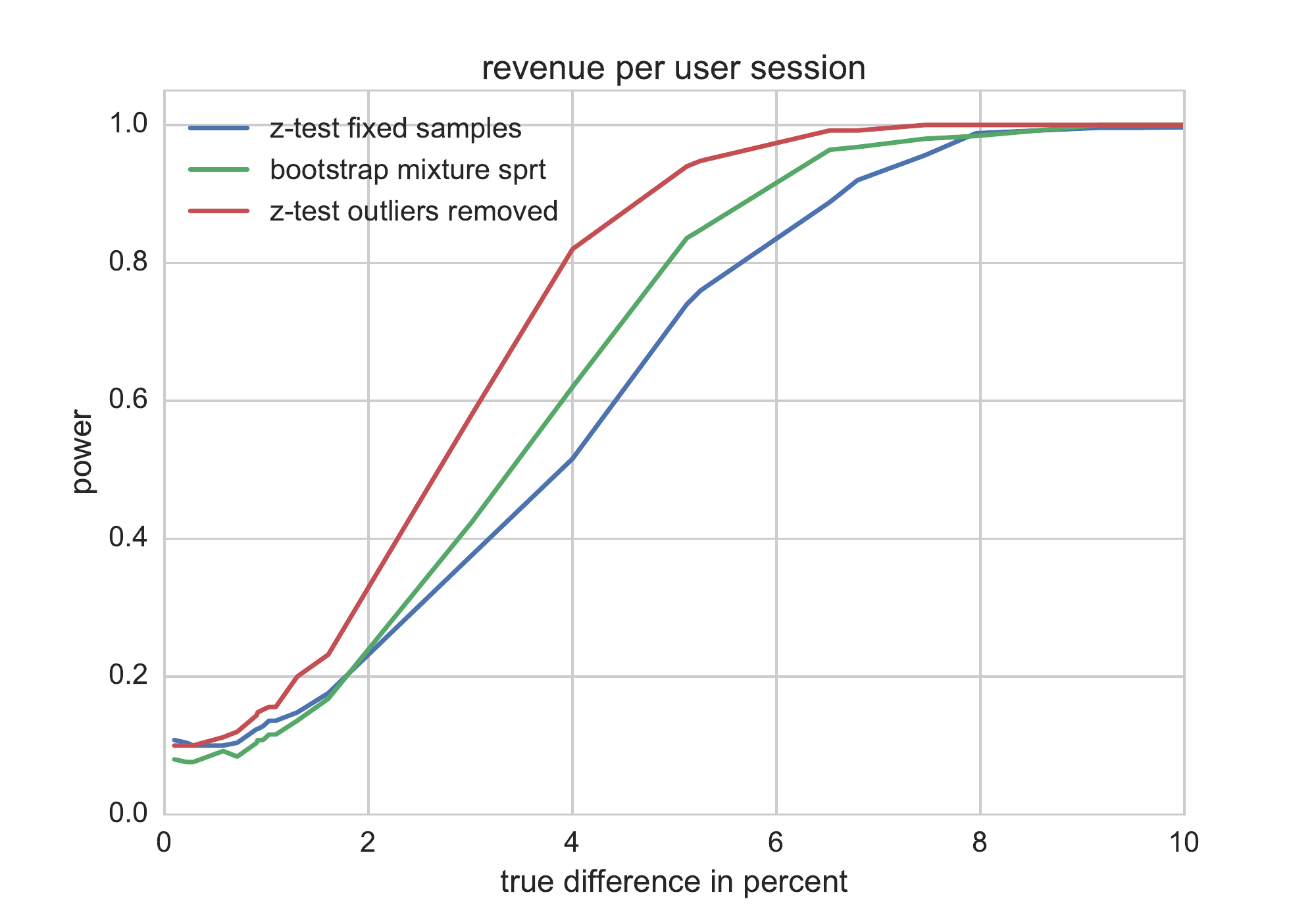}
\caption{\small \sl test power for revenue success rate per user session.\label{fig:revenue-power}} 
\end{subfigure}
\end{center}
\caption{Power comparison for z-test with fixed sample size and the bootstrap mixture SPRT.}
\end{figure}

\textbf{Test power and average duration}:
Next, we compare the power of the bootstrap mixture SPRT with the z-test on the entire data. We use the delta method to compute the standard deviation for the query success rate; see \cite{kempen-2000}. We use an additive offset to the realized difference, as described earlier in this section. Figure \ref{fig:atc-power} shows the power of the test as a function of the true difference percentage for the case of query success rate. Here the true difference percentage is the additive offset normalized by the value of the metric on the entire dataset. Notice some loss in power compared to the z-test on the entire dataset. This loss in power is expected because of the sequential nature of the procedure on a finite data set; equivalently, the bootstrap mixture SPRT requires a larger number of samples to achieve the same power as the fixed sample size tests. This is the price paid for the flexibility of continuous monitoring. The loss in power quickly becomes negligible as the true difference becomes large. Focusing on the revenue metric, we notice something surprising: the z-test on the entire data has smaller power than the bootstrap mixture SPRT, as shown in Figure \ref{fig:revenue-power}. In this case, the mean of data does not confirm to the normal distribution used to justify the z-test. This is illustrated by the z-test on the entire data after outliers removal which has higher power than the bootstrap mixture SPRT, as expected from a test with an apriori access to the entire data.\footnote{Identifying outliers is a complicated problem by itself, often with no good agreement on the definition of outliers. Here, we simply clip the values to some upper limit affecting only a negligible portion of data to show the effect of extreme values in data on the z-test.} The robustness of the bootstrap mixture SPRT to outliers comes from two factors: (i) the use of the bootstrap for learning the likelihood which typically works better for the heavy tail distributions; (ii) working with blocks of data confines the effect of extreme values in the data in a block to that block only by making the likelihood ratio for that block close to one without affecting the other blocks. Thus, not only the bootstrap mixture SPRT allows continuous monitoring, it is more robust to the misspecification in the distribution generating the data. 

\begin{figure}[ht] 
\begin{center}
\includegraphics[trim=0.0in 0.0in 0.65in 0.25in, clip=true, height=2.1in]{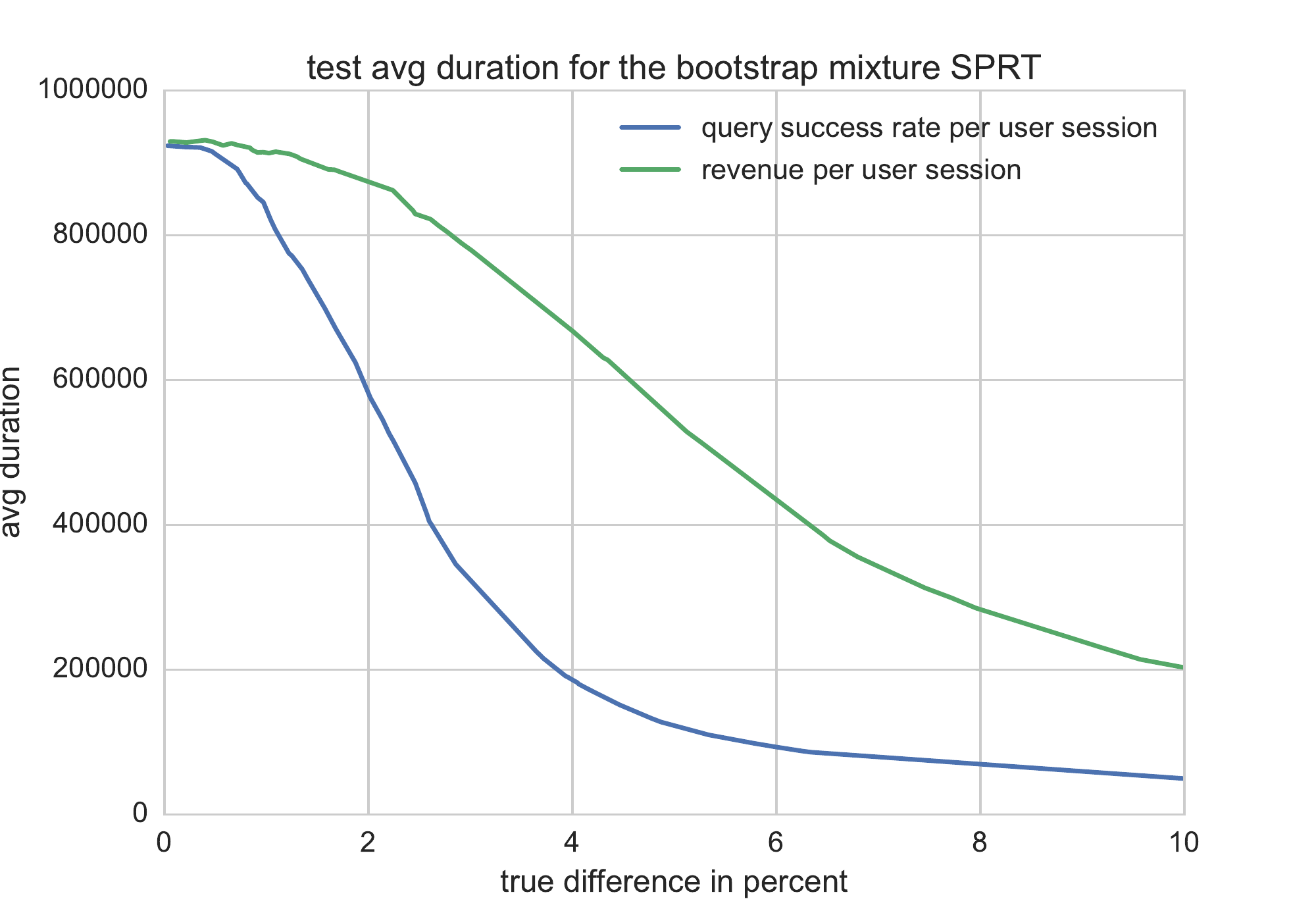}
\caption{\small \sl Test average duration of the bootstrap mixture SPRT for different metrics.\label{fig:1689-duration}} 
\end{center} 
\end{figure}

Figure \ref{fig:1689-duration} shows the average duration of the test using the bootstrap SPRT. The test duration is the numbers of samples consumed in each split until either all the data is exhausted or the null hypothesis is rejected. We take the average over the $250$ random splits of the data set. Notice that the null hypothesis is rejected quickly for large changes, unlike the fixed sample size test where one has to wait until the precomputed number of samples have been collected. This avoids wasteful data collection and unnecessary wait in case of misspecification in the minimum effect size used to determine the number of samples to be collected for fixed sample size tests. 

In our experiments, we observed the power of the tests with smaller block size to be either comparable or better than the power of the tests with larger block size; this is in-line with the observations made earlier about the advantage of the small block size. Similarly, the tests with smaller block size had smaller test duration.

\begin{figure}[t] 
\begin{center}
\begin{subfigure}[t]{0.5\textwidth}
\centering
\includegraphics[trim=0.25in 0.0in 0.65in 0.25in, clip=true, height=2in]{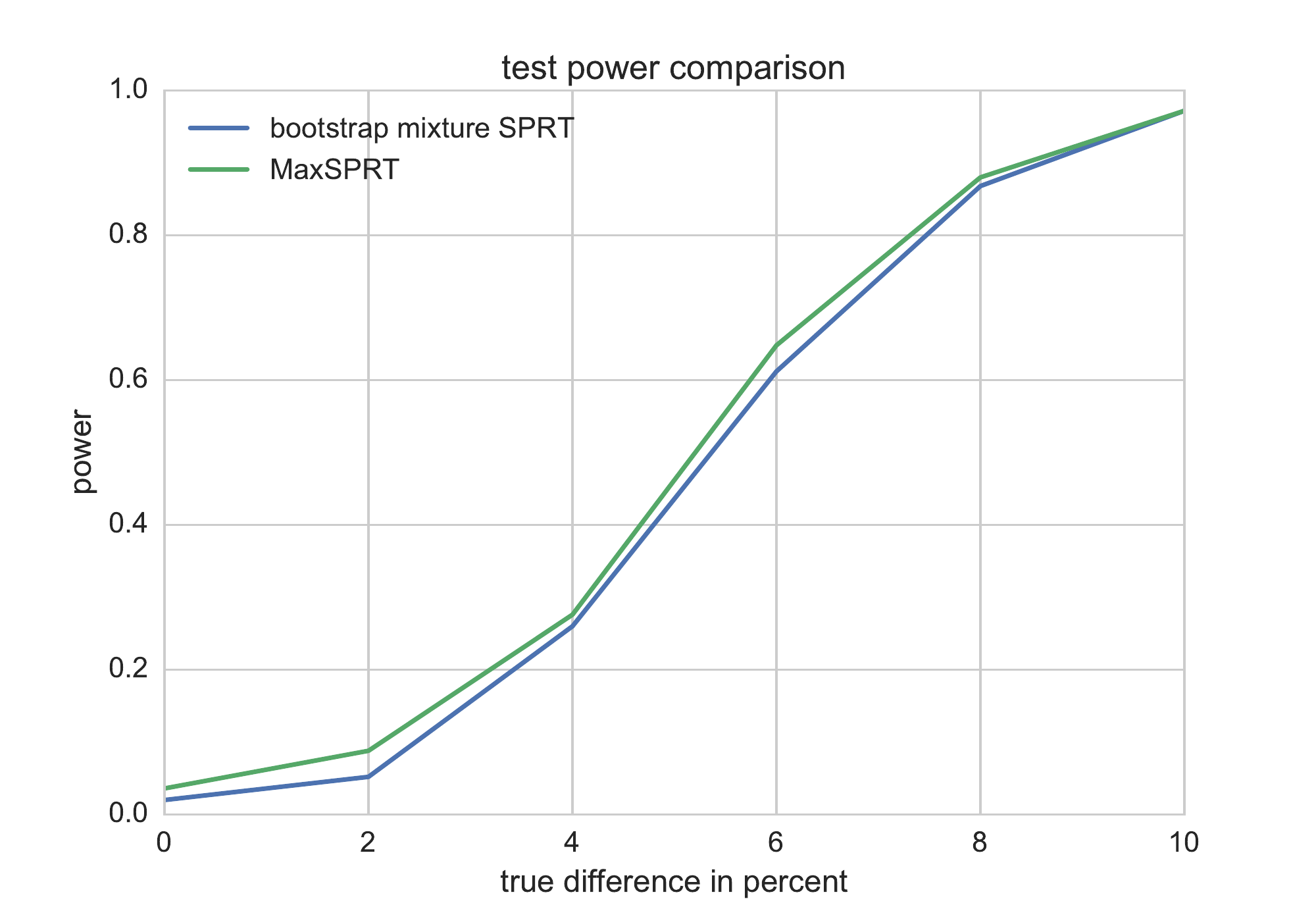}
\caption{\small \sl Test power for the bootstrap mixture SPRT and MaxSPRT.\label{fig:maxsprt-power}} 
\end{subfigure}~~
\begin{subfigure}[t] {0.5\textwidth}
\centering
\includegraphics[trim=0.25in 0.0in 0.65in 0.25in, clip=true, height=2in]{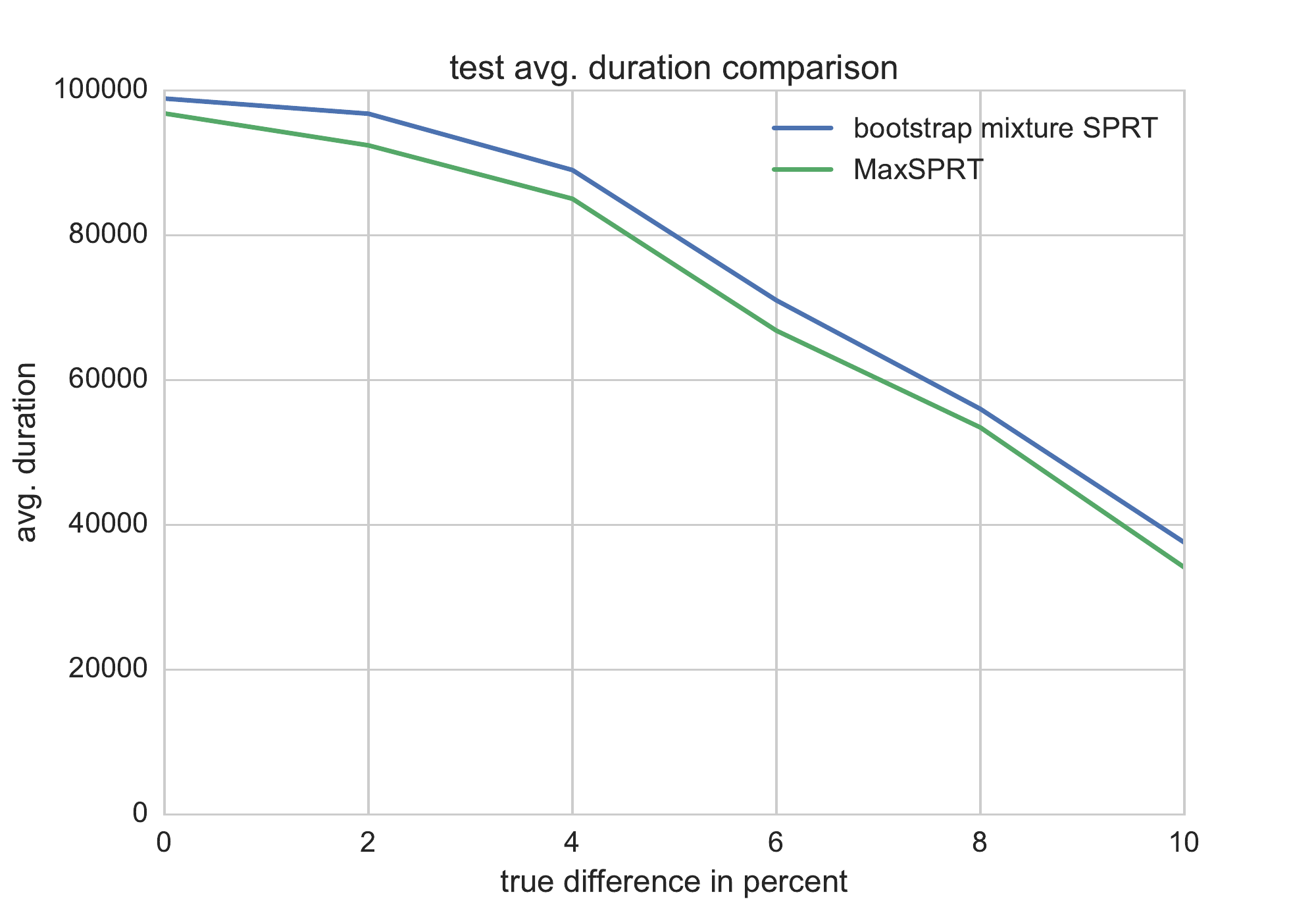}
\caption{\small \sl Average test duration for the bootstrap mixture SPRT and MaxSPRT.\label{fig:maxsprt-duration}} 
\end{subfigure}
\end{center}
\caption{Comparison of the bootstrap mixture SPRT with MaxSPRT.}
\end{figure}

\textbf{Comparison with MaxSPRT on synthetic data}:
Finally, we compare the performance of the bootstrap mixture SPRT with MaxSPRT. We use synthetic data of size $100,000$ generated from the Bernoulli distribution with success probability equal to $0.05$. The rejection threshold for MaxSPRT is determined by multiple A/A tests on synthetic datasets such that the type $1$ error is close to $0.05$, as described in \cite{kharitonov-2015}. We also use A/A tests to determine the block size and the variance of the Gaussian prior for the bootstrap SPRT such that the type $1$ is close to $0.05$. This resulted in the block size of 1000. Figure \ref{fig:maxsprt-power} shows comparable test power for MaxSPRT and the bootstrap mixture SPRT with MaxSPRT performing marginally better. The better performance of MaxSPRT is expected; here the data distribution is exactly known and is exploited by MaxSPRT. The bootstrap mixture SPRT, however, learns the distribution in a nonparametric way and yet performs comparable to MaxSPRT. A similar observation can be made for the average test duration: MaxSPRT has slightly smaller test duration as shown in Figure \ref{fig:maxsprt-duration}. MaxSPRT updates its likelihood ratio score for each data point as it arrives. By contrast the bootstrap mixture SPRT updates its likelihood ratio score in blocks of data which contributes to a slightly longer average test duration. Block update, however, is a more realistic scenario in a practical system where data usually arrives in blocks of size few hundred to a few thousand. The general purpose nature of the bootstrap mixture SPRT while still performing comparable to MaxSPRT which requires exact knowledge of the data distribution is appealing.

\section{Conclusions} \label{sec:conc}
We propose the bootstrap mixture SPRT for the hypothesis test of complex metrics in a sequential manner. We highlight the theoretical results justifying our approach and provide empirical evidence demonstrating its virtues using real data from an e-commerce website. The bootstrap mixture SPRT allows the flexibility of continuous monitoring while controlling the type $1$ error at any time, has good power, is robust to misspecification in the distribution generating the data, and allows quick inference in case the true change is larger than the estimated minimum effect size. The nonparametric nature of this test makes it suitable for the cases where the data generation distribution is unknown or hard to model, which typically is the case.

\bibliographystyle{unsrt}
\bibliography{/Users/vabhish/work/reports/reference/misc}

\end{document}